\documentclass[journal]{IEEEtran}
\ifCLASSINFOpdf
\else
\fi
\usepackage{graphicx, subfig}
\usepackage{epstopdf}
\usepackage{amsmath,bm}
\usepackage[numbers,sort&compress]{natbib}
\usepackage{makecell}
\usepackage{multirow}
\usepackage{color}
\usepackage{multicol}
\usepackage{booktabs}
 \usepackage{amssymb}
 \usepackage{setspace}
\usepackage{fancyhdr}

\begin{document}
%
\title{This work has been submitted to the IEEE for possible publication. Copyright may be transferred without notice, after which this version may no longer be accessible. We will make the paper open access to make sure it can be accessible.  TPPO: A Novel Trajectory Predictor with Pseudo Oracle}


\author{Biao~Yang, \textit{Member}, \textit{IEEE}, Caizhen~He,  Pin~Wang, \textit{Member}, \textit{IEEE}, Ching-Yao~Chan, \textit{Member}, \textit{IEEE}, Xiaofeng~Liu\textsuperscript{*}, \textit{Member}, \textit{IEEE},   Yang~Chen, \textit{Member}, \textit{IEEE}
\thanks{B. Yang, C. Zhen and Y. Chen  are with the Department of Information Science and Engineering, Changzhou University, Changzhou, 213000 China.}
\thanks{P. Wang and C. Chan are with the California PATH, University of California, Berkeley, Richmond, CA, 94804, USA.}
\thanks{B. Yang and X. Liu is with the College of IoT Engineering, Hohai University, Changzhou, 213000, China.}
\thanks{Corresponding author: X. Liu (xfliu@hhu.edu.cn)}}

%



\IEEEtitleabstractindextext{%
\begin{abstract}
Forecasting pedestrian trajectories in dynamic scenes remains a critical problem in various applications, such as autonomous driving and socially aware robots. Such forecasting is challenging due to human-human and human-object interactions and future uncertainties caused by human randomness. Generative model-based methods handle future uncertainties by sampling a latent variable. However, few studies explored the generation of the latent variable. In this work, we propose the \textbf{\emph{T}}rajectory \textbf{\emph{P}}redictor with \textbf{\emph{P}}seudo \textbf{\emph{O}}racle (TPPO), which is a generative model-based trajectory predictor. The first pseudo oracle is pedestrians' moving directions, and the second one is the latent variable estimated from ground truth trajectories. A social attention module is used to aggregate neighbors' interactions based on the correlation between pedestrians' moving directions and future trajectories. This correlation is inspired by the fact that pedestrians' future trajectories are often influenced by pedestrians in front. A latent variable predictor is proposed to estimate latent variable distributions from observed and ground-truth trajectories. Moreover, the gap between these two distributions is minimized during training. Therefore, the latent variable predictor can estimate the latent variable from observed trajectories to approximate that estimated from ground-truth trajectories. We compare the performance of TPPO with related methods on several public datasets. Results demonstrate that TPPO outperforms state-of-the-art methods with low average and final displacement errors. The ablation study shows that the prediction performance will not dramatically decrease as sampling times decline during tests.
\end{abstract}

}

\maketitle

\IEEEdisplaynontitleabstractindextext

%
\IEEEpeerreviewmaketitle

\section{Introduction}
Forecasting pedestrian trajectories in dynamic scenes remains a critical problem with various applications, such as autonomous driving \cite{hong2019rules}, social behavior-aware robots \cite{luber2010people}, and intelligent tracking system \cite{chen2019detecting}. For example, an autonomous driving car could plan a safer route in the crossings if pedestrians' future trajectories in the sidewalks could be accurately predicted by the roadside equipment, such as cameras on buildings or roads. Fig. 1 shows a scenario of trajectory prediction. Pedestrians' future trajectories labeled with different colored arrows are predicted based on past trajectories marked with different colored lines. A human observer can most likely forecast that the woman (blue) with a shoulder bag will turn right to avoid the still car. However, a robot without prior knowledge or solid training may not do the same thing since it is hard to understand human-human and human-object interactions. Several inherent human properties, including interpersonal, socially acceptable, and multi-modal properties, also pose critical challenges for robots to perform accurate trajectory prediction.

\begin{figure}[!h]
\centering
\includegraphics[width=0.5\textwidth]{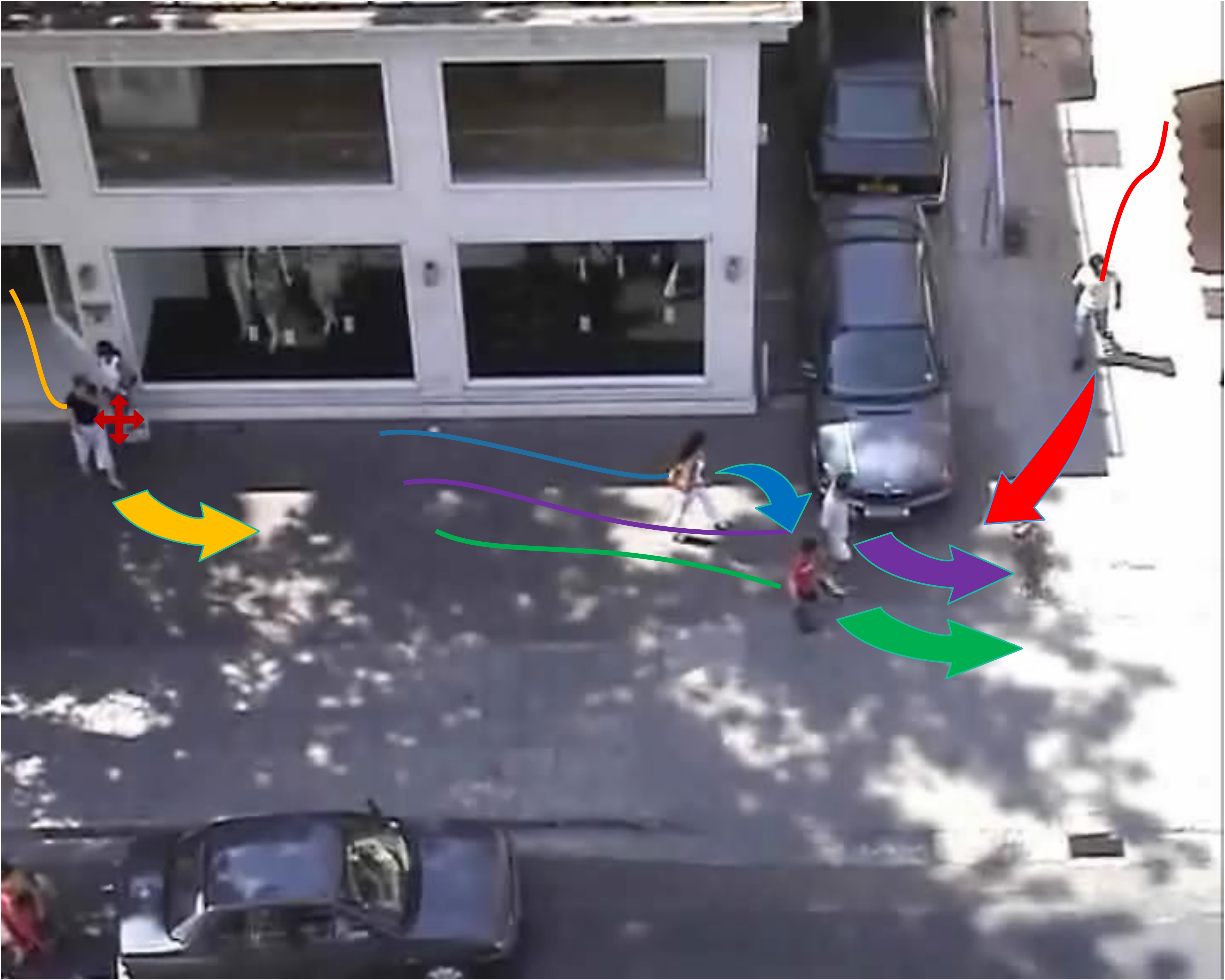}
\caption{Trajectory prediction. For pedestrians in the scene, solid lines are observed trajectories, and the task is to predict their future trajectories marked as arrows. A cross symbol represents a still pedestrian. Humans can easily complete the task. But intelligent machines, such as robots, cannot easily do so (best viewed in color).}
\label{fig:1}
\end{figure}

Early works related to trajectory prediction always depend on simple motion models, such as constant velocity and acceleration models. However, trajectory prediction could not be performed in complex scenes. Later, researchers utilize several stochastic models (e.g., Gaussian mixture regression \cite{li2018generic} or dynamic Bayesian networks (DBNs) \cite{kasper2012object}) to model complex motion patterns. However, such models need hand-crafted features, and thus are difficult to implement. Recent trends in deep learning, especially the recurrent neural network (RNN), provide a data-driven manner to understand complex motion patterns. Specifically, the long-short term memory (LSTM) network is used to encode pedestrians' motion patterns, and their future trajectories are estimated by sampling from the hidden states \cite{alahi2016social}. Recently, a social generative adversarial network (SGAN) is proposed to generate socially acceptable and multi-modal trajectories \cite{gupta2018social}. Except for the understanding of motion patterns, human-human interactions are modeled by a social pooling module based on the common-sense rules. Human-object interactions are captured by feeding the model with the scene-level and object-level information through object detection and semantic segmentation. A detailed review is discussed in the next section.

Although remarkable progress has been made over the past years, there remain two shortcomings that influence trajectory prediction performance. First, few works utilize the correlation between pedestrians' head orientations and their future trajectories. Common sense is that a pedestrian's future trajectory is always influenced by pedestrians in front. Utilizing such common sense is beneficial to model pedestrians' social interactions. Second, few studies explore the latent variable, which plays a critical role in generating multi-modal predictions. Most works use random Gaussian noise as a latent variable. However, such latent variable encompasses little knowledge about the scene or pedestrians and thus is not a proper choice to improve trajectory prediction.

In this work, we propose \textbf{\emph{T}}rajectory \textbf{\emph{P}}redictor with \textbf{\emph{P}}seudo \textbf{\emph{O}}racle (TPPO), which adversarially trains a generator and discriminator. The generator contains an LSTM-based encoder-decoder and the discriminator has an LSTM-based encoder. As Hasan et al. \cite{hasan2018seeing} indicated, pedestrians' head orientations can be used as an oracle for an improved trajectory prediction. We propose to use pedestrians' moving directions as a pseudo oracle, which approximately reflects their head orientations. Then, we utilize the correlation between pedestrians' moving directions and their future trajectories through an attention mechanism. Such a mechanism highlights the influences among correlated pedestrians and thus can better model pedestrians' social interactions. Besides, the latent variable distribution of ground truth trajectories can be regarded as another oracle. We propose a novel latent variable predictor, which estimates two latent variable distributions from observed and ground truth trajectories, respectively. Then, their Kullback-Leibler divergence (KL-divergence) is minimized during training. In the testing stage, the predictor estimates the latent variable distribution of observed trajectories. Such a distribution is similar to that of ground truth trajectories and thus can be regarded as another pseudo oracle. Pedestrians' positions, velocities, and accelerations are extracted from their trajectories and then are fed into the latent variable predictor. Positions reflect the potential scene layout, whereas velocities and accelerations represent pedestrians' motion patterns and radicalness. Notably, the proposed latent variable predictor learns knowledge from trajectories only, and thus increases little computational overhead.

Generally, our contributions are three-fold. (1) We propose a social attention pooling module that fully utilizes the correlation between pedestrians' moving directions and future trajectories. The attention mechanism improves the modeled social interactions. (2) We propose a novel latent variable predictor that can estimate a knowledge-rich latent variable for an improved prediction performance. Such a latent variable is learned from trajectories and thus increase little computational overhead. (3) We embed the social attention pooling module and the latent variable predictor into a GAN framework to generate socially acceptable and multi-modal outputs. Moreover, we achieve state-of-the-art performance on ETH \cite{pellegrini2010improving} and UCY \cite{leal2014learning} datasets.

The rest of the paper is organized as follows. Section \uppercase\expandafter{\romannumeral2} reviews related works. Section \uppercase\expandafter{\romannumeral3} describes the proposed method in detail. Section
\uppercase\expandafter{\romannumeral4} presents the experimental results. Section
\uppercase\expandafter{\romannumeral5} provides the conclusion and discussion.

\section{Related work}
\subsection{Trajectory prediction methods}
Trajectory prediction is a modeling problem that attempts to understand pedestrians' motion patterns by examining pedestrian time-series data. Early works often focus on predicting future trajectories with dynamics-based methods, including constant velocity and acceleration models \cite{zernetsch2016trajectory}. However, a simple kinematic model is not suitable for long-term predictions. For long prediction horizons, flow-based methods \cite{zhi2019spatiotemporal}\cite{molina2018modelling} are proposed to learn the directional flow from observed trajectories in the scene. Subsequently, trajectories are generated by recursively sampling the distribution of future motion derived from the learned directional flow. To cope with complicated scenarios, researchers have resorted to several learning-based methods, such as Gaussian mixture regression \cite{li2018generic}, Gaussian process \cite{laugier2011probabilistic}, random tree searching \cite{aoude2011mobile}, hidden Markov models \cite{wang2018learning}, and DBNs \cite{kasper2012object}. However, it is nontrivial to handle high-dimensional data with these traditional methods.

With the rise of deep learning, RNN provides a data-driven manner of encoding pedestrians' motion series. Prediction models have two categories, namely, deterministic and generative models. For deterministic models, the distribution of future trajectories is estimated from the hidden state encoded by an LSTM or a gated recurrent unit (GRU). Alahi et al. \cite{alahi2016social} presented Social-LSTM, which models pedestrian motions with a shared LSTM and then performs trajectory prediction through sampling. Zhang et al. \cite{zhang2019sr} proposed SR-LSTM, which recursively refines the hidden states of LSTM, to recognize the critical current intention of neighbors. Refined hidden states guarantee improved prediction performance. Some works attempt to embed additional information. Xue et al. \cite{xue2018ss} and Syed et al. \cite{syed2019sseg} utilized three LSTMs to encode person, social, and scene scale information and then aggregate them for context-aware trajectory prediction. Ridel et al. \cite{ridel2019scene} presented a joint representation of the scene and past trajectories by using Conv-LSTM and LSTM, respectively. Lisotto et al. \cite{lisotto2019social} improved Social-LSTM by encompassing prior knowledge about the scene as a navigation map that embodies most frequently crossed areas. Moreover, the scene context is obtained through semantic segmentation to restrain motion for additional plausible paths. These methods improve the prediction performance of Social-LSTM. In contrast to LSTM-based trajectory predictions, several relevant studies attempted to the Transformer architecture \cite{devlin2018bert}, which  introduces a well-designed attention mechanism to encode historical trajectories. Except for forecasting trajectories from the bird's eye view \cite{giuliari2021transformer}\cite{yuan2021agentformer}, these Transformer-based methods can predict pedestrians' trajectories from the first perspective \cite{ngiam2021scene} and 3D scenes \cite{li2020end}. However, they suffer from future uncertainties, which pose a significant challenge in trajectory prediction.

Generative models handle future uncertainties by introducing an alterable latent variable. Lee et al. \cite{lee2017desire} presented DESIRE, which generates multi-modal predictions with conditional variational auto-encoders. Later, Gupta et al. \cite{gupta2018social} proposed SGAN, which adversarially trains the generator and discriminator to produce socially acceptable trajectories. SGAN used the random Gaussian noise as the latent variable and thus generated diverse outputs. Zhu et al. \cite{zhu2019starnet} proposed StarNet, which is similar to SGAN except for the use of a query module. Amirian et al. \cite{amirian2019social} replaced the L2 loss used in SGAN with the information loss \cite{chen2016infogan} to avoid mode collapse.

Besides methods mentioned above, reinforcement learning-based trajectory predictors  \cite{ma2017forecasting}\cite{li2019way}\cite{van2019safecritic}\cite{deo2019scene}\cite{fernando2019neighbourhood} exhibit a rising trend. Moreover, reinforcement learning-based methods always provide the optimal trajectory, whereas the true trajectory is sub-optimal due to pedestrians' randomness. However, estimating pedestrians' destinations that are essential for reinforcement learning is difficult due to future uncertainties.

\subsection{Social interaction modeling}
Great efforts in trajectory prediction are devoted to modeling pedestrians' social interactions. These interactions can be defined by hand-crafted rules, such as social forces \cite{helbing1995social} and stationary crowds' influences \cite{yi2015understanding}. However, capturing complex interactions in the scene with these rules is difficult. Therefore, current works always learn social interactions in a data-driven manner. Social-LSTM \cite{alahi2016social} first proposed the social pooling layer to aggregate neighbors' information in a particular grid. Pei et al. \cite{yi2015understanding} presented a social-affinity map by replacing the grid with a bin. SGAN \cite{gupta2018social} removed the regional restraint and directly aggregated all neighbors' information in the scene. To further capture pedestrians' social interactions, Amirian et al. \cite{amirian2019social} replaced the relative displacement with the bearing angle, the Euclidean distance, and the closest distance between the target person and neighbors. These additional variables calculated from observed trajectories are more intuitional in modeling social interactions than relative displacements. They also used an attention mechanism, which is useful in highlighting key neighbors and features as proposed in SoPhie \cite{sadeghian2019sophie}.

Except for the social pooling layer and its variants, graph neural network is another mainstream to model pedestrians' social interactions \cite{kosaraju2019social} \cite{huang2019stgat}. Such methods are especially suitable for handling heterogeneous agents \cite{ma2019trafficpredict} or crowd scenarios \cite{ivanovic2019trajectron}. However, both methods neglect the correlation between pedestrian's head orientation and trajectory prediction. As Hasan et al. \cite{hasan2018seeing} reported, knowing the head orientation is beneficial to modeling social interactions. Specifically, future trajectories of the target person are influenced by pedestrians in front. However, such an oracle cue is difficult to estimate from image data. In this work, we approximately take pedestrians' moving directions as their head orientations and model their social interactions with a social attention pooling module.

\subsection{Latent variable learning}
Generative model-based prediction methods have gone mainstream due to their ability to handle future uncertainties. The latent variable used in the generator has a strong correlation with generated multi-modal outputs. Several previous works \cite{gupta2018social}\cite{sadeghian2019sophie} used random Gaussian noise as the latent variable and injected little knowledge about pedestrians or scenes into the generator. Amirian et al. \cite{amirian2019social} sampled part of the latent variable from a uniform distribution on the interval from 0 to 1. However, associating the trajectory or the scene with a specific latent variable remains difficult. Zhang et al. \cite{zhang2019stochastic} presented a stochastic module to generate the latent variable based on pedestrians' movements. The latent variable is sampled from the embedding outputs of a social graph network. Tang et al. \cite{tang2019multiple} proposed a dynamic encoder to learn latent variables from multiple inputs, including trajectories and the environmental context. However, processing the visual context needs more computation than processing trajectory data alone.

In this work, we propose a novel predictor that only learns the latent variable from trajectory data. Specifically, we feed pedestrians' positions, velocities, and accelerations into the predictor to understand the latent environmental context, pedestrians' motion patterns and radicalness. Unlike References \cite{zhang2019stochastic} \cite{tang2019multiple} those only utilized observed data, we attempt to minimize the latent variable distribution gaps between observed and ground truth trajectories. Our inspiration comes from Reference \cite{denton2018stochastic}, which focused on stochastic video generation with a learned prior.

\section{Proposed method}
In this work, we develop a trajectory predictor that can generate multiple future trajectories of all agents in a scene with high accuracy. Fig. 2 illustrates the system pipeline of TPPO.

\begin{figure*}
\centering
\includegraphics[width=1\textwidth]{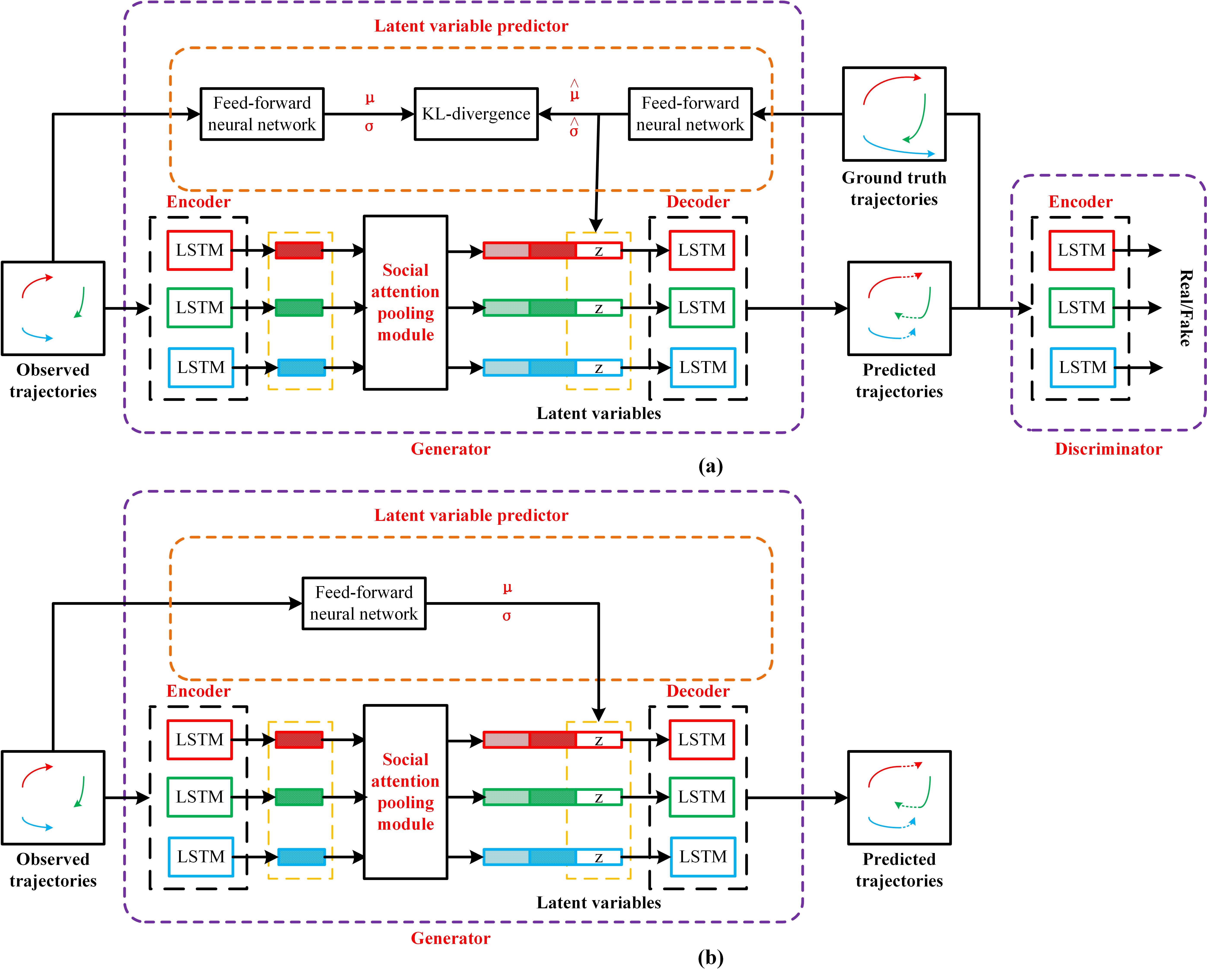}
\caption{System pipeline. Our model contains three key components: a generator, a discriminator, and a latent variable predictor. In the (a) training stage, the generator and discriminator are trained in an adversarial manner. The generator first encodes the input trajectories with an LSTM-based encoder, and then interactions are calculated by the social attention pooling module. Outputs of the encoder and pooling module are concatenated together with the latent variable estimated from ground-truth trajectories. An LSTM-based decoder then decodes these outputs to predict trajectories. The discriminator takes the input ground-truth and predicted trajectories and classifies them as socially acceptable or not. Only the generator is reserved in the (b) testing stage, and the latent variable is estimated from observed trajectories (best viewed in color).}
\label{fig:2}
\end{figure*}

\subsection{Problem definition}
The trajectory prediction problem is a time-series analysis. For pedestrian \emph{i}, we first denote the position ($x_{i}^{t}$, $y_{i}^{t}$) at time step \emph{t} as $p_{i}^{t}$. The goal of trajectory prediction is to estimate the future trajectory $\mathcal{T}_{i}=\left(\boldsymbol{p}_{i}^{t+1}, \ldots, \boldsymbol{p}_{i}^{t+\boldsymbol{T}_{obs}}\right)$, considering his motion history $\mathcal{H}_{i}=\left(p_{i}^{0}, \ldots, p_{i}^{t}\right)$ and interactions with other pedestrians or objects. Then, the trajectory prediction problem is converted into training a parametric model that predicts future trajectory $\mathcal{T}_{i}$ ($\emph{i}$=1,...,$\emph{n}$), which can be formulated as follows:

\begin{equation}
\underset{\Theta}{\arg \max } P_{\theta}\left(\mathcal{T}_{1}, \ldots, \mathcal{T}_{n} | \mathcal{H}_{1}, \ldots, \mathcal{H}_{n}\right),
\end{equation}
where $\Theta$ represents learnable parameters, and $\emph{n}$ represents the number of pedestrians. Recently, the formulation mentioned above is always converted into a sequence-to-sequence prediction problem, which the data-driven RNN module can addres.

\subsection{Generator and discriminator}
As reviewed above, generative model-based methods are always used for trajectory prediction due to their ability to generate multi-modal outputs. Therefore, we design the structure of TPPO based on SGAN, which can produce multi-modal and socially acceptable trajectories through adversarial training. We briefly introduce the generator and the discriminator as follows:

$\textbf{Generator}$: The generator consists of a shared encoder-decoder and a social attention pooling module used to model pedestrians' social interactions. A linear layer is used to convert the relative displacements \cite{gupta2018social}\cite{huang2019stgat}\cite{amirian2019social} of pedestrian \emph{i} ($\Delta x_{i}^{t}=x_{i}^{t}-x_{i}^{t-1}$, $\Delta y_{i}^{t}=y_{i}^{t}-y_{i}^{t-1}$) into a fixed-length vector $e_{i}^{t}$. Then, the vector is fed into an LSTM-based encoder to generate the hidden state of pedestrian \emph{i} at time \emph{t} as follows:
\begin{equation}
\begin{aligned} e_{i}^{t} &=\phi\left(\Delta x_{i}^{t}, \Delta y_{i}^{t} ; W_{e e}\right) \\ h_{e i}^{t} &=L S T M\left(h_{e i}^{t-1}, e_{i}^{t} ; W_{e n c o d e r}\right), \end{aligned}
\end{equation}
where $\phi(\cdot)$ is a linear layer function with ReLU nonlinearity. $W_{e e}$ and $W_{e n c o d e r}$ are the learnable weights of $\phi(\cdot)$ and the encoder function $LSTM(\cdot)$, respectively.

Afterward, pedestrians' interactions are aggregated by the social attention pooling module and latent variables are estimated by the latent variable predictor. We will discuss these two modules in detail later. Then, an LSTM-based decoder is used to estimate the future trajectories of pedestrian \emph{i} based on the concatenation of the hidden state $h_{e i}^{t}$, the pooling output $P_{i}^{t}$, and the estimated latent variable $z_{i}^{t}$. The decoding is formulated as follows:
\begin{equation}
\begin{aligned} h_{d i}^{t} &=L S T M\left(h_{d i}^{t-1}, [h_{e i}^{t}, P_{i}^{t}, z_{i}^{t}] ; W_{\text {decoder}}\right) \\\left(\hat{x}_{i}^{t}, \hat{y}_{i}^{t}\right) &=\gamma\left(h_{d i}^{t}\right) \end{aligned},
\end{equation}
where $\gamma(\cdot)$ is a linear layer function with ReLU nonlinearity to predict future trajectories $\left(\hat{x}_{i}^{t}, \hat{y}_{i}^{t}\right)$. $W_{\text {decoder}}$ represents the learnable weight of the decoder, which is shared by all pedestrians in the scene.

$\textbf{Discriminator}$: The discriminator is used to classify the predicted and ground truth trajectories as socially acceptable or not. Both trajectories are fed into an LSTM-based encoder to generate the embedding, and then a Softmax classifier is used to perform classification based on the embedding. Pedestrians' interactions are not needed in the discriminator.

\subsection{Social attention pooling module}
The social pooling layer proposed in SGAN can aggregate pedestrians' interactions in crowded scenes. As shown in Fig. 3, a multilayer perceptron (MLP) module is used to encode relative positions between target A and all other pedestrians. These embedding vectors are concatenated with each pedestrian's hidden state. Another MLP is used to process the concatenated vectors and the results are pooled element-wise to compute target A's pooling vector.

However, such a pooling layer only considers relative positions between target people and their neighbors. As common knowledge, pedestrians' future trajectories are always influenced by people in front. As demonstrated in Fig. 3, target A's future trajectories are influenced by targets B and C, who are in A's field of view (FoV). Target D does not interfere A's trajectory decision even when he runs to A. Several previous works utilized pedestrians' head orientations to infer the FoV \cite{hasan2018seeing}. However, precisely recognizing pedestrians' head orientations from vision data is challenging. Thus, using such common knowledge for an improved trajectory prediction is difficult.

TPPO uses an attention mechanism to utilize the correlation between pedestrians' head orientations and their trajectories. We approximately take pedestrians' moving directions at the last step as their head orientations. Then, the cosine values of all pedestrians' bearing angles are calculated as follows:
\begin{equation}
\cos (\mathcal{B})=\left[\begin{array}{ccc}
{\cos \left(b_{1 1}\right)} & {\cdots} & {\cos \left(b_{1 n}\right)} \\
{\vdots} & {\ddots} & {\vdots} \\
{\cos \left(b_{n 1}\right)} & {\cdots} & {\cos \left(b_{n n}\right)}
\end{array}\right],
\end{equation}
where n is the number of pedestrians in a scene. $b_{ij}$ represents the bearing angle of agent j from agent i, that is, the angle between the velocity of agent i and the vector joining agents i and j.

Afterward, the attentional weights are calculated based on the cosine values. We perform hard and soft attention operations to refine the second MLP's outputs. Finally, outputs of the attention module are max-pooled to generate pedestrian A's pooling vector $P_{A}$. The hard and soft attention operations are formulated as follows:

$\textbf{Hard attention}$: As discussed above, one pedestrian's influence on another decreases when the bearing angle increases. Therefore, hard attention weight is represented as a matrix $H_{A}$ of the same size of $cos (\mathcal{B})$, and each element $h_{ij}$ in $H_{A}$ is set to 0 or 1 by thresholding. $h_{ij}$ is set to 1 if $cos(b_{ij})$ is larger than an empirically threshold $-0.2$, otherwise is set to 0.

$\textbf{Soft attention}$: Unlike hard attention that calculates attention weight by thresholding, soft attention adaptively calculates various pedestrian pairs' correlations. Soft attention weight $S_{A}$ can be formulated as follows:
\begin{equation}
S_{A}=\delta(\varphi(\cos (\mathcal{B}))),
\end{equation}
where $\delta(\cdot)$ represents the sigmoid activation, and $\varphi(\cdot)$ represents the $1 \times 1$ convolution.

\begin{figure*}
\centering
\includegraphics[width=1\textwidth]{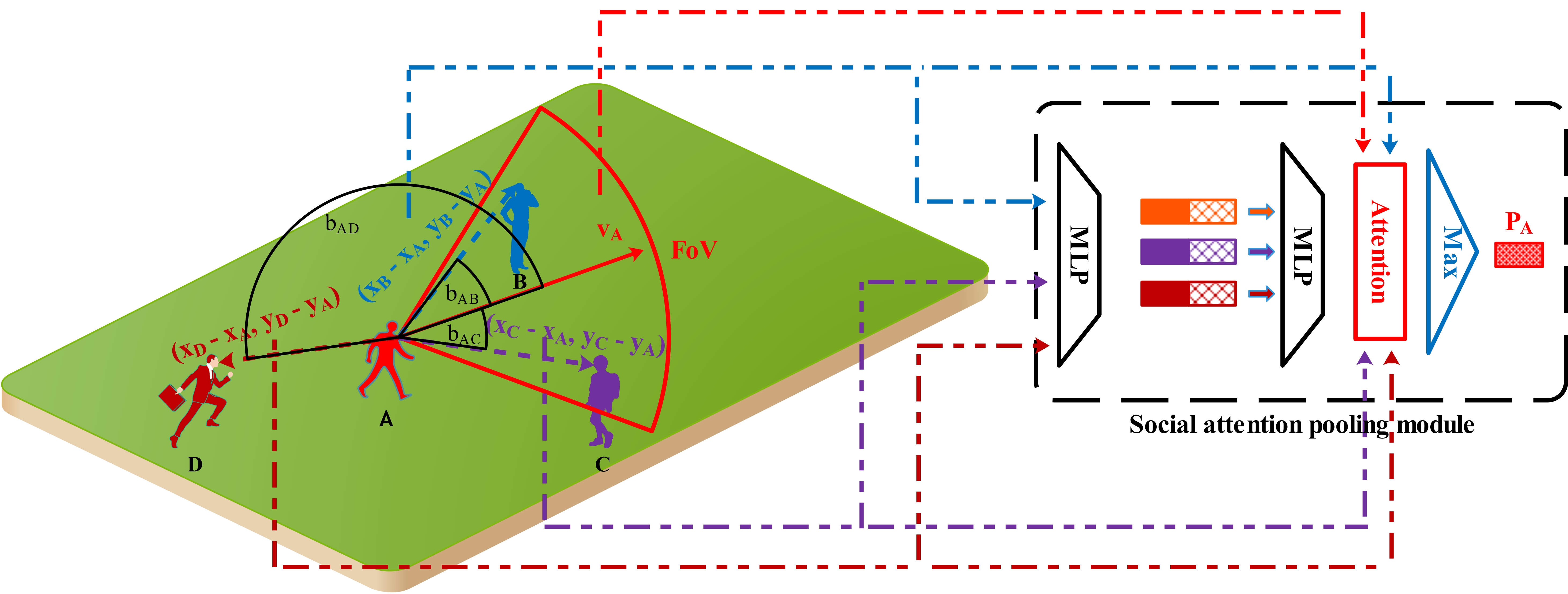}
\caption{Social attention pooling module. Compared with the social pooling layer proposed in SGAN, we add an attention module with the soft or hard operation. The left part illustrates the social effects between pedestrian A and his neighbors. FoV is defined based on A's velocity $V_{A}$. Intuitively, pedestrians B and C in the FoV can influence A's future trajectories, whereas D contributes little to A's trajectory prediction. The right part demonstrates the pipeline of information pooling. Outputs of the second MLP are weighted by the attention mechanism and then are pooled element-wise to compute A's pooling vector $V_{A}$. The proposed social attention pooling module can model interactions between all pairs of pedestrians (best viewed in color).}
\label{fig:3}
\end{figure*}

\subsection{Latent variable predictor}
The latent variable predictor is used to estimate an alterable latent variable for generating multi-modal outputs. Specifically, we attempt to train a well-learned predictor, which can estimate similar latent variable distributions from observed and ground truth trajectories, respectively. Afterward, the predictor can estimate knowledge-rich latent variables from observed trajectories only in the testing stage. Inputs to the latent variable predictor are pedestrians' positions, velocities, and accelerations. As shown in Fig. 2, the latent variable predictor consists of two feed-forward neural networks, which are formulated as follows:
\begin{equation}
\begin{array}{l}{\left(\mu_{i}^{k}, \sigma_{i}^{k}\right)=\Psi\left(I_{i}^{k} ; W_{L P}^{k}\right)} \\ {\left(\hat{\mu}_{i}^{k}, \hat{\sigma}_{i}^{k}\right)=\hat{\Psi}\left(\hat{I}_{i}^{k} ; \hat{W}_{L P}^{k}\right)}\end{array},
\end{equation}
where $\Psi(\cdot)$ and $\hat{\Psi}(\cdot)$ are the feed-forward neural networks with learnable weights $W_{L P}^{k}$ and $\hat{W}_{L P}^{k}$, respectively. $I_{i}^{k}$ and $\hat{I}_{i}^{k}$ are the $k^{th}$ kinds of inputs (positions, velocities, and accelerations) we extract from observed and ground-truth trajectories, respectively. $\left(\mu_{i}^{k}, \sigma_{i}^{k}\right)$ and $\left(\hat{\mu}_{i}^{k}, \hat{\sigma}_{i}^{k}\right)$ represent the latent variable distributions of the $k^{th}$ kind of input estimated by $\Psi(\cdot)$ and $\hat{\Psi}(\cdot)$, respectively. The final latent variable z is generated by concatenating the latent variable distributions of three kinds of inputs and the random Gaussian noise. $\left(\hat{\mu}_{i}^{k}, \hat{\sigma}_{i}^{k}\right)$ and $\left(\mu_{i}^{k}, \sigma_{i}^{k}\right)$ are used in the training and testing stages, respectively. The random Gaussian noise is used to generate multi-modal outputs as that used in SGAN. The latent variable estimated by the proposed predictor is expected to provide rich information for precise trajectory prediction while maintaining the ability to handle future uncertainties.

\subsection{Loss function}
The loss function used in this work consists of three parts: adversarial, variety, and latent variable distribution losses. The adversarial loss is used to generate socially acceptable trajectories by performing a two-player min-max game. Unlike traditional GAN, the latent variable $\emph{z}$ used in this work is sampled from the ground truth trajectory at the training stage with the latent variable predictor. Thus, the adversarial loss is formulated as follows:
\begin{equation}
\begin{array}{l}{\mathcal{L}_{a d v}=} \\ {\mathbb{E}_{x \sim p_{\text {data}}(x)}[\log D(x)]+\mathbb{E}_{z \sim p_{(z | \hat{I}_{i}^{k})}}[\log (1-D(G(z | \hat{I}_{i}^{k})))]}\end{array},
\end{equation}

The variety loss is used to fit the best-predicted trajectory in L2 loss while maintaining multi-modal outputs. We follow its definition proposed in SGAN, and the variety loss is defined as follows:
\begin{equation}
\mathcal{L}_{\text {variety}}=\min _{m}\left\|\hat{\mathcal{T}_{i}}-\mathcal{T}_{i}^{m}\right\|_{2},
\end{equation}
where $\hat{\mathcal{T}_{i}}$ and $\mathcal{T}_{i}^{m}$ are ground truth and the $m^{th}$ predicted trajectories, respectively. m is a hyper-parameter and is set to 20 according to SGAN.

The latent variable distribution loss is used to measure the latent variable distribution gaps between observed and ground truth trajectories. We introduce KL-divergence to calculate the loss, which is formulated as follows:
\begin{equation}
\mathcal{L}_{\text {LD}}=D_{K L}(\left(\mu_{i}^{k}, \sigma_{i}^{k}\right) || \left(\hat{\mu}_{i}^{k}, \hat{\sigma}_{i}^{k}\right)),
\end{equation}

Afterward, the total loss is defined in a weighted manner as follows:
\begin{equation}
\mathcal{L}_{\text {total}}=\mathcal{L}_{a d v}+\alpha \times \mathcal{L}_{\text {variety}}+\beta \times \mathcal{L}_{\text {LD}},
\end{equation}
where $\alpha$ and $\beta$ are set to 1 and 10 respectively by cross validation across benchmarking datasets.

\subsection{Implementation details}
LSTMs with 32-dimensional hidden states are used in the encoder and the decoder. We use two three-layer MLPs (2 $\times$ 32 $\times$ 32 for the first one, and 64 $\times$ 32 $\times$ 16 for the second one) in the social attention pooling module. The 16-dimensional latent variable z contains three four-dimensional vectors that embedded from positions, velocities, and accelerations and another four-dimensional random Gaussian noise. We iteratively train the generator and discriminator with a batch size of 64 for 600 epochs using Adam \cite{kingma2014adam}, with the initial learning rate of 0.001. The training frequency of the discriminator is twice that of the generator for better convergence. The proposed model is built with a Pytorch framework and is trained with an NVIDIA GTX-1080 GPU.

\section{Experimental results}
The proposed method is evaluated on two publicly available datasets, namely ETH \cite{pellegrini2010improving} and UCY \cite{leal2014learning}. Both of them consist of real-world pedestrian trajectories with rich human-human and human-object interaction scenarios. There are five sets of data, namely ETH, HOTEL, UNIV, ZARA1, and ZARA2. In total, there are 1,536 pedestrians in challenging scenarios, such as people crossing each other, group forming and dispersing, and collision avoidance. All the trajectories are converted to real-world coordinates, sampled every 0.4 seconds through the interpolation operation.

Similar to several prior works \cite{gupta2018social}\cite{amirian2019social}, the proposed method is evaluated with two error metrics as follows:

1. \emph{Average Displacement Error (ADE)}: Average L2 distance between the ground-truth trajectory and the predicted trajectory over all predicted time steps.

2. \emph{Final Displacement Error (FDE)}: The Euclidean distance between the actual final destination and the predicted final destination at the last step of prediction.

We use the leave-one-out approach for evaluation \cite{alahi2016social}. Specifically, we train models on four sets and test them on the remaining set. The observed and predicted horizons are 8 (3.2 seconds) and 8 / 12 (3.2 / 4.8 seconds) time steps, respectively. We denote $\textbf{\emph{T}}$ as the prediction horizon.

\subsection{Quantitative  evaluations}
\textbf{Comparisons with baseline methods}: We compare the proposed method against the following baselines:

1. \emph{Linear}: A linear regressor is used to predict future trajectories by minimizing the least square error.

2. \emph{LSTM}: An LSTM is used to embed the motion patterns of observed trajectories. Future trajectories are predicted based on the learned motion patterns.

3. \emph{Social-LSTM} \cite{alahi2016social}: An improved LSTM-based trajectory prediction method by proposing a social pooling layer to aggregate hidden states of interested pedestrians. Future trajectories are predicted by decoding the concatenation of LSTM embedding and social pooling outputs.

4. \emph{SGAN} \cite{gupta2018social}: An improved version of Social-LSTM by utilizing adversarial training to generate socially acceptable trajectories. Gaussian noises are used as latent variables to generate multi-modal outputs in consideration of pedestrians' future uncertainties. The model is trained using a variety loss with the hyper-parameter set to 20. During the test, 20 times are sampled from the generator, and the best prediction in L2 sense is used for quantitative evaluation.

Fig. 4 demonstrates the quantitative comparisons between TPPO, with either soft or hard attention modules, and the aforementioned baseline methods across five sets. It is obviously seen that the proposed TPPO outperforms all baseline methods in ADE and FDE for both prediction horizons. We owe the superiority of TPPO to the proposed social attention pooling module and the novel latent variable predictor. Furthermore, TPPO with hard attention module is slightly better than TPPO with soft attention module. Such a difference may indicate that an empirically defined threshold is more suitable to be used in the social attention pooling module since it is challenging to learn a well-trained soft attention function with limited trajectory data.

\begin{figure*}
\centering
\includegraphics[width=0.85\textwidth]{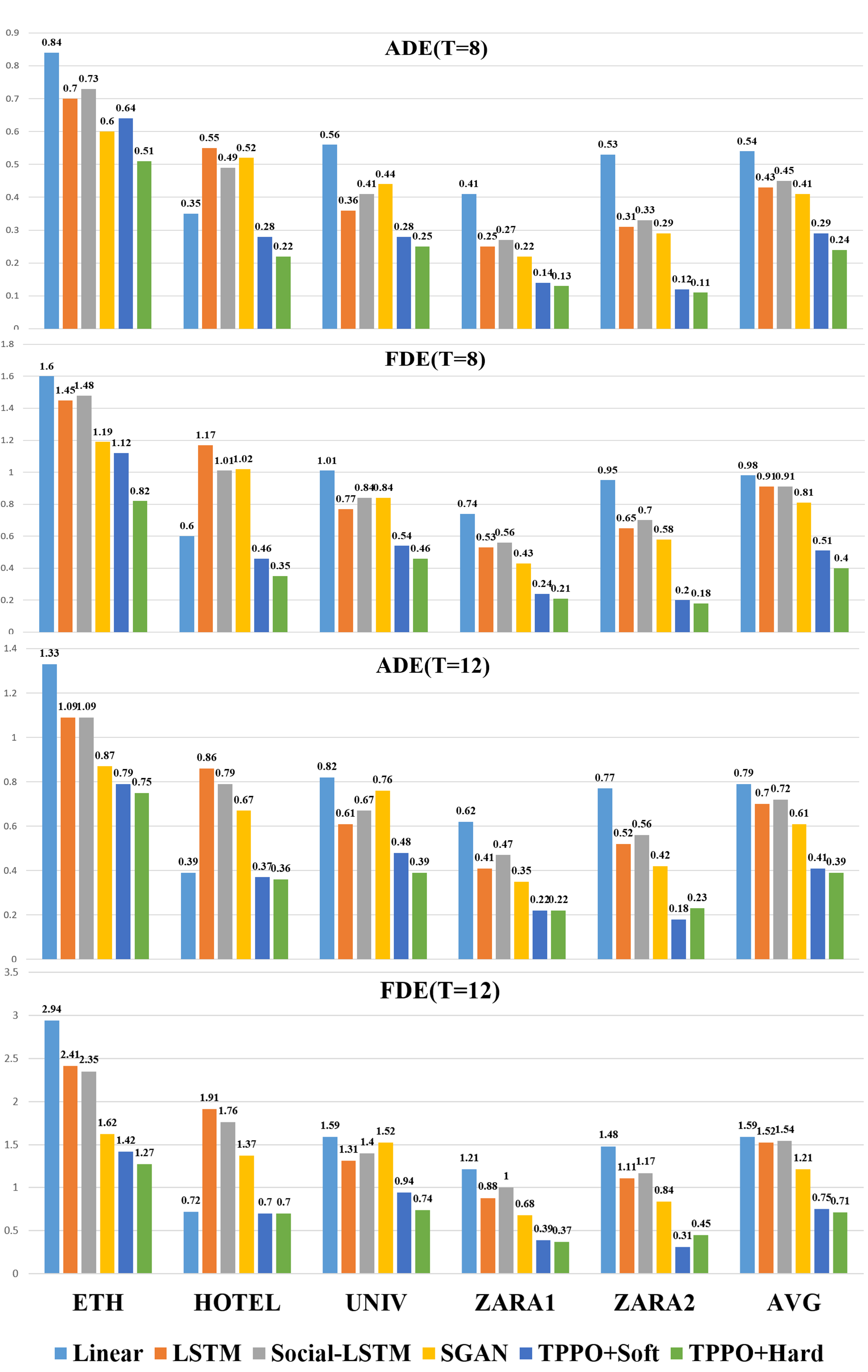}
\caption{Quantitative results between the proposed method and baseline methods mentioned above across five sets. We report the ADE and FDE for $\textbf{\emph{T}}$ = 8 and $\textbf{\emph{T}}$ = 12 (8 / 12) in meters. Our method outperforms other baseline methods (low is preferred and is labeled with bold fonts).}
\label{fig:4}
\end{figure*}

\textbf{Comparisons with state-of-the-art methods}: We compare the proposed method against the following state-of-the-art methods:

1. \emph{SR-LSTM} \cite{zhang2019sr}: An improved version of Social-LSTM by proposing a data-driven state refinement module. Such a module iteratively refines the current pedestrians' hidden states based on their neighbors' intentions through the message passing.

2. \emph{Sophie} \cite{sadeghian2019sophie}: An improved version of SGAN by utilizing attention mechanisms, namely, the social and physical attention modules. The trajectory prediction performance is enhanced by highlighting the critical information with attention operations.

3. \emph{S-Way} \cite{amirian2019social}: An improved version of SGAN by replacing the L2-loss with the information-loss proposed in Reference \cite{chen2016infogan} to avoid mode collapsing.

4. \emph{Social-BiGAT} \cite{kosaraju2019social}: An improved version of SGAN by using the bicycle structure to train the generator. A graph attention network is used to model social interactions for better prediction performance.

5. \emph{STGAT} \cite{huang2019stgat}: An auto-encoder based trajectory prediction method that uses a spatio-temporal graph attention network to model pedestrians' social interactions in the scene. Specifically, the spatial interactions are captured by the graph attention mechanism, and temporal correlations are modeled by a shared LSTM.

6. \emph{CGNS} \cite{li2019conditional}: CGNS forecasts the future trajectories by approximating the data distribution, with which realistic, feasible and diverse future trajectory hypotheses can be sampled. The interaction information is captured with soft attention mechanisms.

7. \emph{AC-VRNN} \cite{bertugli2021ac}: AC-VRNN uses a generative architecture for multi-future trajectory predictions based on conditional variational recurrent neural networks. Human interactions are modeled with a graph-based attention mechanism enabling an online attentive hidden state refinement of the recurrent estimation.

Table \uppercase\expandafter{\romannumeral1} presents the comparison between TPPO and state-of-the-art methods. Similar to results in Fig. 4, TPPO with the hard attention module outperforms others in both prediction horizons in UNIV, ZARA1, and ZARA2 sets. We also report the relative ADE and FDE gains of TPPO with the hard attention module compared with the STGAT, which is the best method among methods selected for comparison. As shown in the column ``Rel. gain'', our method achieves appreciable gains except in ETH and HOTEL sets. Such satisfactory performance benefits from the proposed social attention pooling module and the novel latent variable predictor. Compared with traditional methods that capture pedestrians' social interactions in a data-driven manner, the social attention pooling module introduces obstacle avoidance experiences into the data-driven learning process to better capture social interactions. Furthermore, the latent variable predictor firstly predicts pedestrians' future intentions from observed trajectories, then forecasts accurate future trajectories based on predicted intentions. Compared with the proposed method, S-Ways performs well in ETH and HOTEL sets by designing an attention-based module that focuses on the bearing angle, the Euclidean distance, and the distance of the closest approach. Such a module can capture human-human and human-object interactions in complex scenes that contain lots of obstacles. However, S-Ways performs poorly in ZARA1 and ZARA2 sets. AC-VRNN achieves the sub-optimal average ADE and FDE because of the usage of the graph-based attention mechanism. It reveals that the graph model is good at capturing social interactions, which are significant for accurate trajectory prediction. The online attentive hidden state refinement strategy used in AC-VRNN makes it advantageous over STGAT  and Social-BiGAT that also use graph attention networks. Compared with the state-of-the-art methods, our method achieves the best performance in average ADE and FDE values. Moreover, the hard attention module performs better than the soft attention module.

\begin{table*}
\label{tab1}
\centering
\caption{ Comparison results with state-of-the-art methods across all datasets. We report the ADE and FDE for $\textbf{\emph{T}}$ = 12 in meters. Our method outperforms state-of-the-art methods in UNIV, ZARA1, and ZARA2 sets, and is specifically good for average ADE and FDE (low is preferred and is labeled with bold fonts). ``Rel. gain" shows the relative ADE and FDE gains of our best model + Hard (marked in red) compared with the latest method AC-VRNN (marked in blue).}
\begin{tabular}{llllllllllll}
\toprule
Metric & Dataset & SR-LSTM & Sophie & S-Ways & Social-BiGAT & CGNS   & STGAT & {\color{blue}AC-VRNN}  & TPPO &    &  Rel. gain \\
\cline{10-11}
       &         &         &        &        &              &     &    &    & + Soft & {\color{red}+ Hard}  & \\
\midrule
ADE    & ETH     & 0.63    & 0.70   & \textbf{0.39}   & 0.69    & 0.62     & 0.65  & {\color{blue}0.61}    & 0.79  & {\color{red}0.75}  & -18.67$\%$\\
       & HOTEL   & 0.37    & 0.76   & 0.39   & 0.49    & 0.70     & 0.35  & {\color{blue}\textbf{0.30}}  & 0.37  & {\color{red}0.36} & -16.67$\%$\\
       & UNIV    & 0.51    & 0.54   & 0.55   & 0.55     & 0.48    & 0.52  & {\color{blue}0.58}  & 0.48  & {\color{red}\textbf{0.39}} & +48.72$\%$\\
       & ZARA1   & 0.41    & 0.30   & 0.44   & 0.30   & 0.32      & 0.34  & {\color{blue}0.34}  & \textbf{0.22}  & {\color{red}\textbf{0.22}} & +54.55$\%$\\
       & ZARA2   & 0.32    & 0.38   & 0.51   & 0.36    & 0.35     & 0.29  & {\color{blue}0.28}  & \textbf{0.18}  & {\color{red}0.23} & +21.74$\%$\\
AVG    &         & 0.45    & 0.54   & 0.46   & 0.48    & 0.49     & 0.43  & {\color{blue}0.42}  & 0.41  & {\color{red}\textbf{0.39}} & +7.69$\%$\\
FDE    & ETH     & 1.25    & 1.43   & \textbf{0.64}   & 1.29    & 1.40     & 1.12  & {\color{blue}1.09}  & 1.42  & {\color{red}1.27} & -14.17$\%$\\
       & HOTEL   & 0.74    & 1.67   & 0.66   & 1.01    & 0.93     & 0.66  & {\color{blue}\textbf{0.55}}  & 0.70  & {\color{red}0.70} & -21.43$\%$\\
       & UNIV    & 1.10    & 1.24   & 1.31   & 1.32    & 1.22     & 1.10  & {\color{blue}1.22}  & 0.94  & {\color{red}\textbf{0.74}} & +64.86$\%$\\
       & ZARA1   & 0.90    & 0.63   & 0.64   & 0.62     & 0.59      & 0.69  & {\color{blue}0.68}  & 0.39  & {\color{red}\textbf{0.37}} & +83.78$\%$\\
       & ZARA2   & 0.70    & 0.78   & 0.92   & 0.75     & 0.71    & 0.60  & {\color{blue}0.59}  & \textbf{0.31}  & {\color{red}0.45} & +31.11$\%$\\
AVG    &         & 0.94    & 1.15   & 0.83   & 1.00     & 0.97     & 0.83  & {\color{blue}0.83}  & 0.75  & {\color{red}\textbf{0.71}} & +16.90$\%$\\
\bottomrule
\end{tabular}
\end{table*}

\subsection{Ablation study}
We conduct an ablation study to demonstrate the effects of various modules used in TPPO. Soft and Hard represent soft and hard attention modules, respectively. The MLVP module represents the latent variable predictor with multiple inputs. For comparison, we propose the SLVP module, which uses the velocity only to estimate the latent variable. As presented in Table \uppercase\expandafter{\romannumeral2}, the SLVP module does not improve the prediction performance of SGAN. However, the MLVP module drastically decreases the ADE and FDE errors in most sets compared with SGAN. Such findings confirm the effectiveness of multiple inputs in latent variable prediction. Furthermore, the improved results verify that the proposed latent variable predictor could accurately predict pedestrians' future trajectories by leveraging  their predicted intentions. The latent variable predictor is our significant contribution compared with other trajectory prediction approaches. Besides, we embed hard and soft attention modules into SGAN to evaluate the social attention pooling layer's effectiveness. For SGAN, the soft attention module is slightly better than the hard attention module, whereas an opposite conclusion is drawn when both attention modules are used in TPPO. The possible reason for such slight nonconformity is the differences in network structures. Specifically, TPPO makes a trade-off between the learning of the soft attention module and latent variable predictor, whereas SGAN only focuses on the soft attention module during training.

\begin{table*}
\label{tab2}
\centering
\caption{ Ablation study. We report the ADE and FDE for $\textbf{\emph{T}}$ = 8 and $\textbf{\emph{T}}$ = 12 (8 / 12) in meters across five sets. SLVP / MLVP represent $\textbf{s}$ingle / $\textbf{m}$ultiple inputs $\textbf{l}$atent $\textbf{v}$ariable $\textbf{p}$redictor, respectively. Soft / Hard represent soft and hard attention mechanisms, respectively (low is preferred and is labeled with bold fonts). }
\begin{tabular}{lllllllll}
\toprule
Metric & Dataset & SGAN & + SLVP & + MLVP & + Soft & + Hard & + MLVP + Soft & + MLVP + Hard \\
\midrule
ADE    & ETH     & 0.60/0.87    & 0.71/0.94   & 0.64/\textbf{0.67}   & 0.73/0.84         & 0.67/0.86  & 0.64/0.79  & \textbf{0.51}/0.75 \\
       & HOTEL   & 0.52/0.67    & 0.42/0.65   & 0.29/0.44   & 0.36/0.49         & 0.50/0.51  & 0.28/0.37  & \textbf{0.22}/\textbf{0.36} \\
       & UNIV    & 0.44/0.76    & 0.40/0.70   & 0.27/0.45   & 0.38/0.63         & 0.39/0.62  & 0.28/0.48  & \textbf{0.25}/\textbf{0.39} \\
       & ZARA1   & 0.22/0.35    & 0.23/0.40   & 0.17/0.24   & 0.21/0.32         & 0.21/0.34  & 0.14/\textbf{0.22}  & \textbf{0.13}/\textbf{0.22} \\
       & ZARA2   & 0.29/0.42    & 0.21/0.34   & 0.13/0.21   & 0.20/0.29         & 0.20/0.29  & 0.12/\textbf{0.18}  & \textbf{0.11}/0.23 \\
AVG    &         & 0.41/0.61    & 0.39/0.61   & 0.30/0.40   & 0.38/0.51         & 0.39/0.52  & 0.29/0.41  & \textbf{0.24}/\textbf{0.39} \\
FDE    & ETH     & 1.19/1.62    & 1.35/1.96   & 1.12/\textbf{1.13}   & 1.37/1.67         & 1.28/1.71  & 1.12/1.42  & \textbf{0.82}/1.27 \\
       & HOTEL   & 1.02/1.37    & 0.81/1.40   & 0.58/0.91   & 0.68/0.98         & 0.96/1.08  & 0.46/\textbf{0.70}  & \textbf{0.35}/\textbf{0.70} \\
       & UNIV    & 0.84/1.52    & 0.83/1.49   & 0.51/\textbf{0.69}   & 0.79/1.34         & 0.81/1.34  & 0.54/0.94  & \textbf{0.46}/0.74 \\
       & ZARA1   & 0.43/0.68    & 0.46/0.82   & 0.30/0.42   & 0.41/0.64         & 0.40/0.68  & 0.24/0.39  & \textbf{0.21}/\textbf{0.37} \\
       & ZARA2   & 0.58/0.84    & 0.43/0.73   & 0.22/0.36   & 0.40/0.60         & 0.40/0.61  & 0.20/\textbf{0.31}  & \textbf{0.18}/0.45 \\
AVG    &         & 0.81/1.21    & 0.78/1.28   & 0.55/\textbf{0.70}   & 0.73/1.05         & 0.77/1.08  & 0.51/0.75  & \textbf{0.40}/0.71 \\
\bottomrule
\end{tabular}
\end{table*}

\subsection{Evaluations of different sampling times}
Despite the necessity of generating multi-modal outputs, a good trajectory predictor should estimate precise future trajectories with few shots. One advantage of TPPO is its ability to forecast precise trajectories with few shots. We perform a comparison between SGAN and TPPO with the MLVP module only by gradually decreasing sampling times. Fig. 5 illustrates the comparison results of ADE and FDE between TPPO and SGAN when using different sampling times across five sets. Except for the ADE and FDE for $\textbf{\emph{T}}$ = 12 in the HOTEL set, TPPO significantly outperforms SGAN, especially with few sampling times. We can observe that the ADE and FDE values of TPPO gently increase while sampling times are decreasing. However, those of SGAN drastically increase when the number of samples is close to 1. The difference indicates the power of TPPO in predicting precise future trajectories with few shots.

\begin{figure*}
\centering
\includegraphics[width=1.0\textwidth]{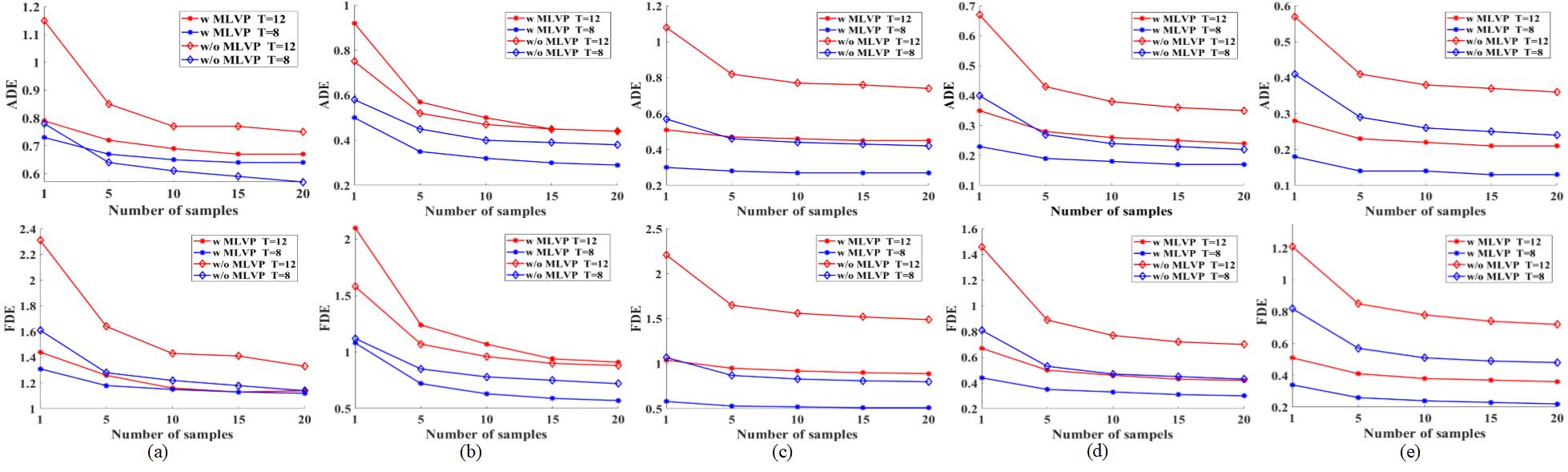}
\caption{Comparison results of ADE and FDE values between TPPO and SGAN when using different numbers of samples across (a) ETH, (b) HOTEL, (c) UNIV, (d) ZARA1, and (e) ZARA2 sets. The upper and lower parts are comparison results of ADE and FDE, respectively. The star marker represents the proposed method, and the diamond marker represents SGAN. The red and blue lines (including the dashed) represent the prediction results for $\textbf{\emph{T}}$ = 12 and $\textbf{\emph{T}}$ = 8 in meters, respectively (best viewed in color and zoom-in.)}
\label{fig:5}
\end{figure*}

\subsection{Qualitative evaluations}
TPPO performs precise trajectory prediction based on the latent variable predictor, which estimates a knowledge-rich latent variable from the trajectory data only. Fig. 6 demonstrates the trajectory prediction results by using SGAN, SR-LSTM, Sophie, and TPPO with different modules in different datasets. Specifically, we show the prediction results of TPPO with the MLVP module only, with MLVP and soft attention modules, and with MLVP and hard attention modules. Each sub-figure represents a scenario with multiple pedestrians. For each pedestrian, the predicted trajectory is the best one with the lowest ADE value among the 20 samples generated by each method. Generally, all methods can predict future trajectories with high accuracy at most of the time. However, TPPO performs better than the selected state-of-the-art methods with predicted future trajectories closer to the ground truth. Moreover, as illustrated in the fourth scenario of Fig. 6(c), TPPO can handle a sudden motion change, which is a challenging issue in trajectory prediction. Therefore, the information learned from positions, velocities, and accelerations is useful for accurate trajectory prediction in most cases. However, such information may mislead TPPO to generate wrong results in rare cases, as shown in the third scenario of Fig. 6(b). We will investigate the robustness of the latent variable predictor in our future work.

\begin{figure*}
\centering
\includegraphics[width=1.0\textwidth]{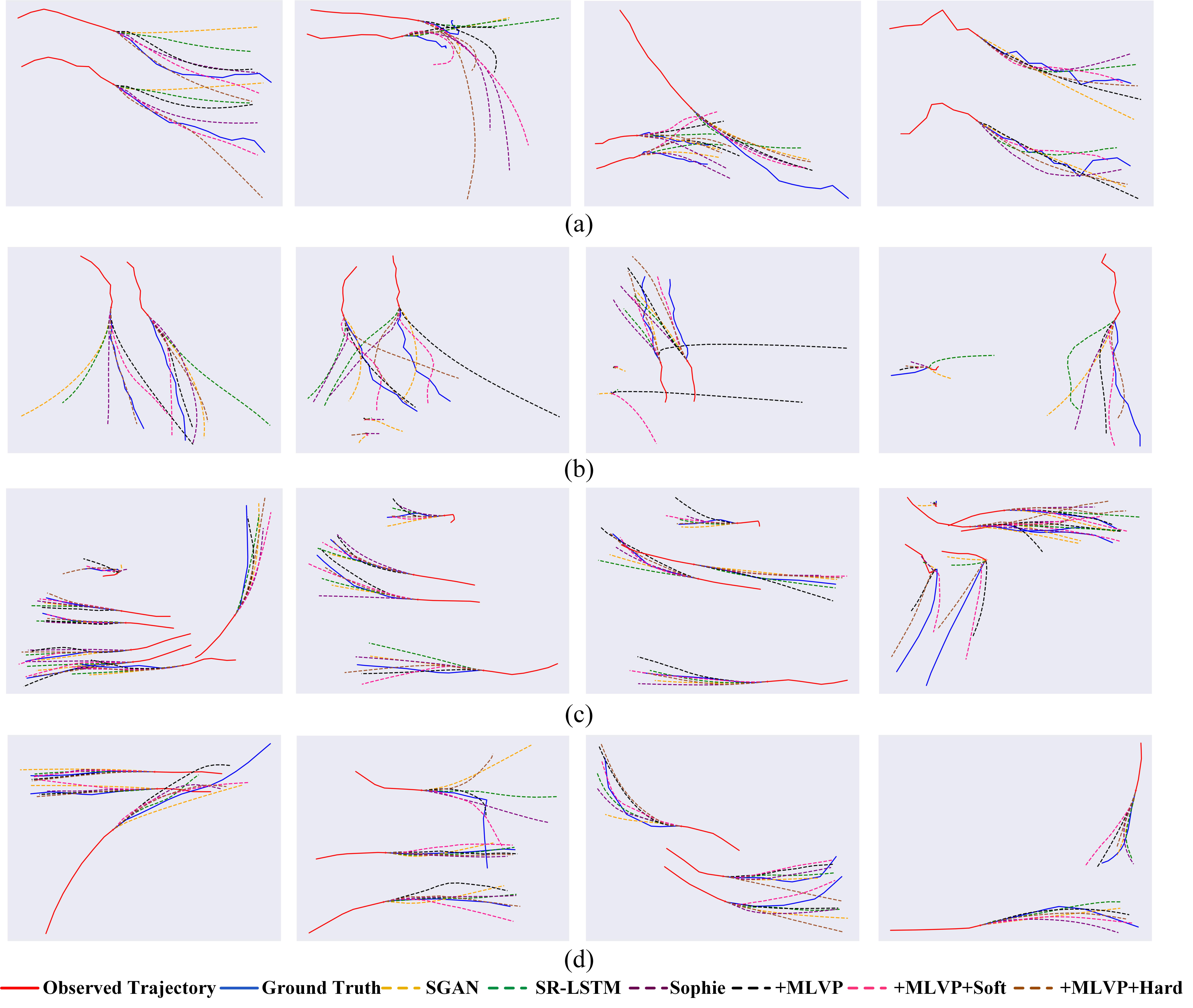}
\caption{Trajectory prediction results using SGAN, SR-LSTM, Sophie, and TPPO with different modules in (a) ETH, (b) HOTEL, (c) ZARA1, and (d) ZARA2 sets. Red and blue lines represent observed and ground-truth trajectories, respectively. Dashed lines of different colors represent the predicted trajectories of different methods. We show the best trajectory with the lowest ADE value from 20 predicted samples (best viewed in color and zoom-in).}
\label{fig:6}
\end{figure*}

TPPO forecasts socially acceptable trajectories while maintaining diverse outputs by injecting the random Gaussian noise into the predicted latent variable. Fig. 7 illustrates the density maps of the predicted trajectories in four typical scenarios randomly selected from different data sets. The typical scenario refers to the scenario contains pedestrians' social interactions, e.g., people crossing each other, group forming and dispersing. Generating density maps in these scenarios exhibits our method's ability to forecast pedestrians' future trajectories and capture pedestrians' social interactions. Density maps in the UNIV set are not shown since too many trajectories are present in that set. In the first row, the MLVP module helps TPPO recognize the change of motion direction, whereas methods without the MLVP module fail. In the second row, the MLVP module helps TPPO avoid the tree, and the attention module encourages the model to generate separate outputs, which are socially acceptable. The two scenarios in ZARA1 and ZARA2 sets reveal the ability of TPPO in handling motion changes. The difference between these two scenarios is that the fourth row demonstrates a crowded scene. Therefore, the attention module can improve the prediction results compared with TPPO using the MLVP module alone.

\begin{figure*}
\centering
\includegraphics[width=1.0\textwidth]{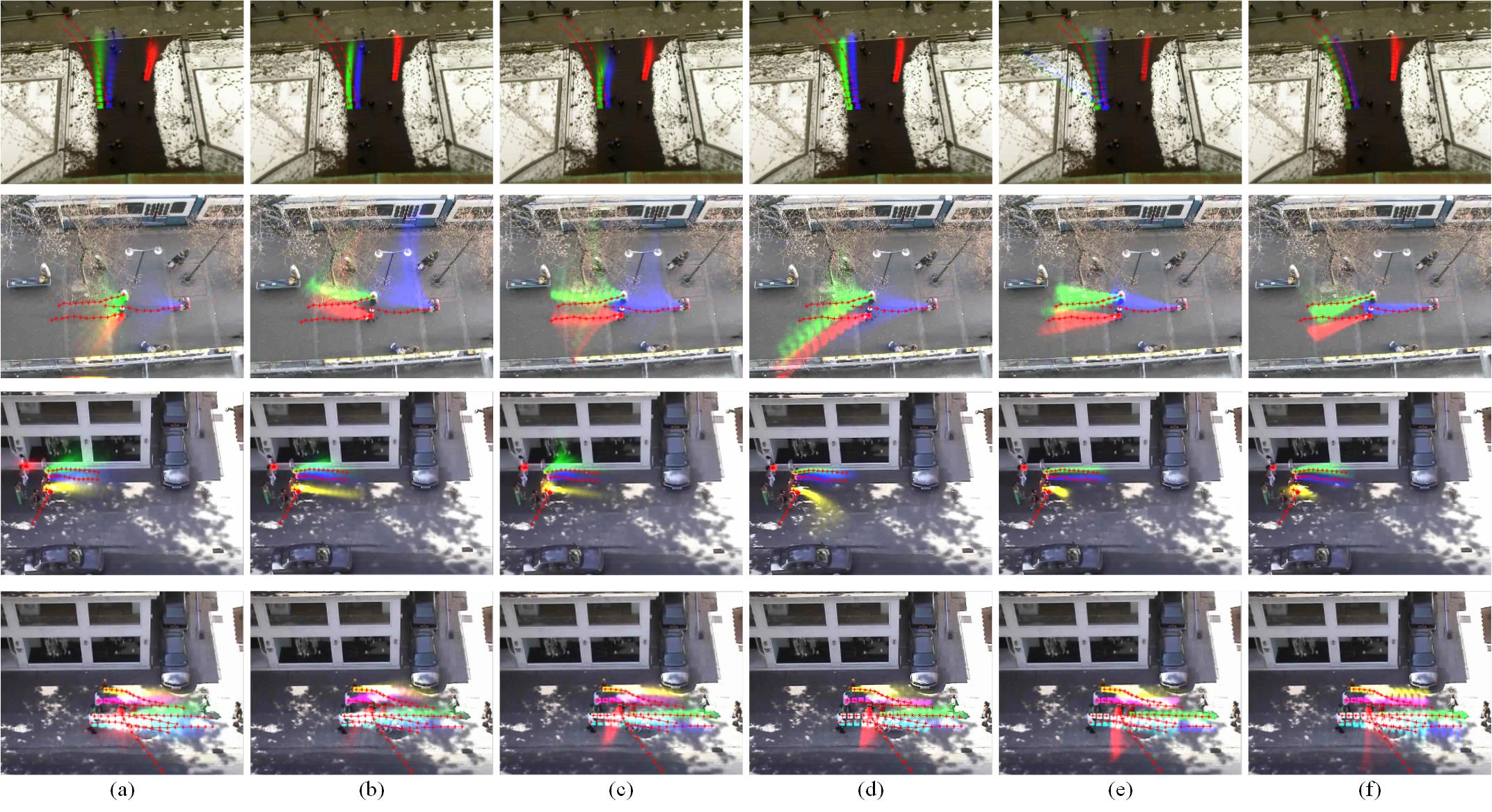}
\caption{Density maps of the predicted trajectories by using SGAN and TPPO with different modules, including (a) SGAN, (b) + Soft, (c) + Hard, (d) + MLVP, (e) + MLVP + Soft, and (f) + MLVP + Hard. The first, second, third, and fourth rows are typical scenarios randomly selected from the ETH, HOTEL, ZARA1, and ZARA2 sets, respectively. The density maps are generated by sampling 300 times from the learned generators. The red stars represent the true future trajectories, and different colors indicate the density distributions of different pedestrians (best viewed in color and zoom-in).}
\label{fig:7}
\end{figure*}

\section{Conclusion and discussion}
In this work, we propose TPPO, which forecasts future trajectories with two pseudo oracles. One pseudo oracle is pedestrians' moving directions, which are used to approximate pedestrians' head orientations. A social attention pooling module utilizes this pseudo oracle to improve trajectory prediction performance. Another pseudo oracle is the latent variable distribution estimated from ground truth trajectories. We propose a novel latent variable predictor, which estimates the latent variable distribution from observed trajectories. Such a distribution is similar to that from ground truth. The random Gaussian noise is injected into the estimated latent variable to handle future uncertainties. Evaluations are performed in two commonly used metrics, namely, ADE and FDE, across two benchmarking datasets. Comparisons with state-of-the-art approaches indicate the effectiveness of the proposed latent variable predictor. Ablation studies reveal the necessity of learning information from multiple inputs and the superiority of accurate trajectory prediction with few sampling times. Besides, the proposed method only learns knowledge from trajectories and thus increases little computing overhead. Our future work focuses on how to control the latent variable for an improved prediction performance and forecast pedestrians' future trajectories from the first perspective. Moreover, the proposed method will be deployed on a self-navigating robot, which can make more informed decisions on selecting an optimal path after sensing agents' future trajectories.

\section*{Acknowledgment}
This study was supported partially by the National Key Research and Development Program 2018AAA0100800; Key Research and Development Program of Jiangsu under grants BK20192004, BE2018004-04; International and Exchanges of Changzhou under grant CZ20200035; Postdoctoral Foundation of Jiangsu Province No.2021K187B; National Postdoctoral General Fund No.2021M701042; and the Changzhou Sci$\&$Tech Program with Grant No.CJ20210052.


%




\ifCLASSOPTIONcaptionsoff
  \newpage
\fi



%


\bibliographystyle{IEEEtran}
\bibliography{TPPO}

%

\begin{IEEEbiography}[{\includegraphics[width=1in,height=1.25in,clip,keepaspectratio]{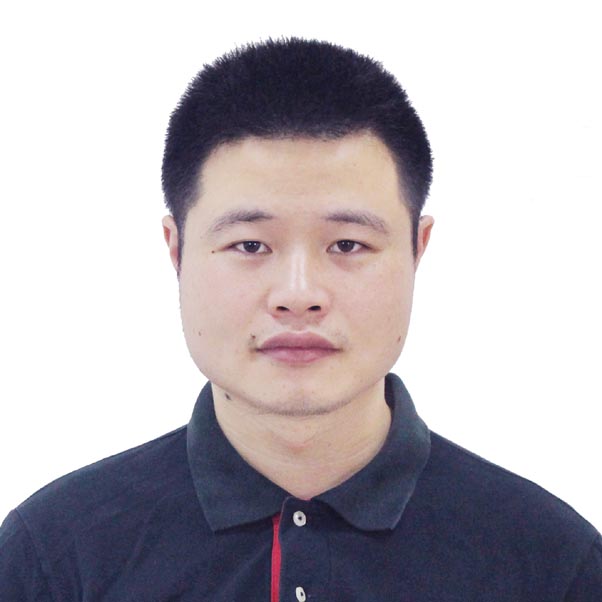}}]{Biao Yang}
received his BS degree from Nanjing University of Technology. He received his MS and Ph.D degrees in instrument science and technology from Southeast University, Nanjing, China, in 2014. From 2018 to 2019 he was a visiting scholar in the University of California, Berkeley. Now he works at Changzhou University, China. His current research interests include machine learning and pattern recognition based on computer vision.
\end{IEEEbiography}
\vspace{-35pt}
\begin{IEEEbiography}[{\includegraphics[width=1in,height=1.25in,clip,keepaspectratio]{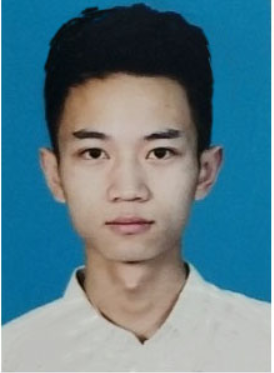}}]{Caizhen He}
was born in Yangzhou, China, in 1996. He received the B.S degree in engineering from Changshu Institute of Technology 2018. He is currently pursuing the master degree in computer science.
\end{IEEEbiography}
\vspace{-35pt}
\begin{IEEEbiography}[{\includegraphics[width=1in,height=1.25in,clip,keepaspectratio]{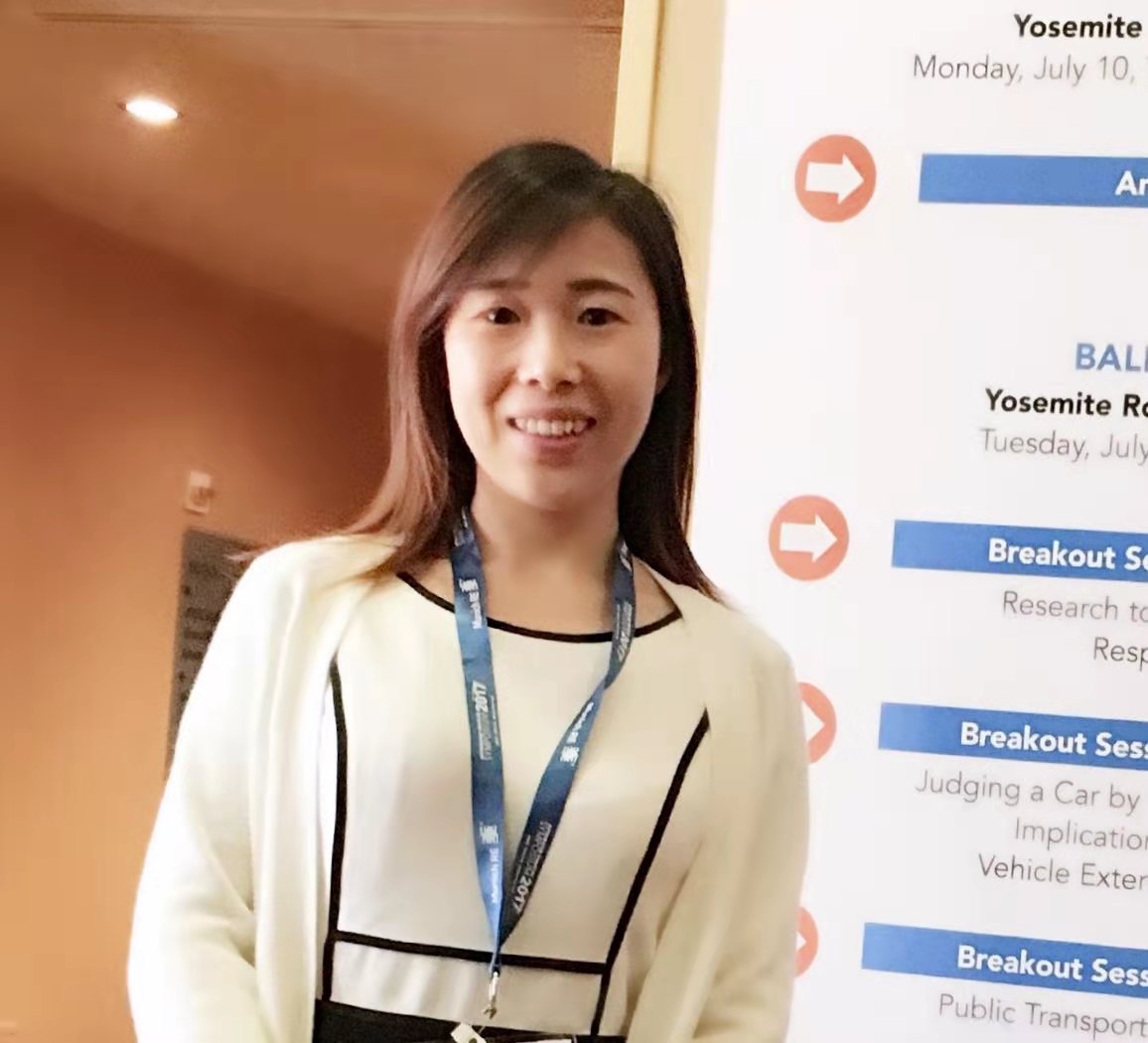}}]{Pin Wang}
received the Ph.D. degree in Transportation Engineering from Tongji University, China, in 2016, and was a postdoc at California PATH, University of California, Berkeley, from 2016 to 2018. She is now is a researcher and team leader at California PATH, UC Berkeley. Her current research is focused on Deep Learning algorithms and applications for Autonomous Driving, including driving behavior learning, trajectory planning, control policy learning, and pedestrian intention prediction. She also collaborates with industries on projects such as intelligent traffic control and advanced vehicular technology assessment.
\end{IEEEbiography}
\vspace{-35pt}
\begin{IEEEbiography}[{\includegraphics[width=1in,height=1.25in,clip,keepaspectratio]{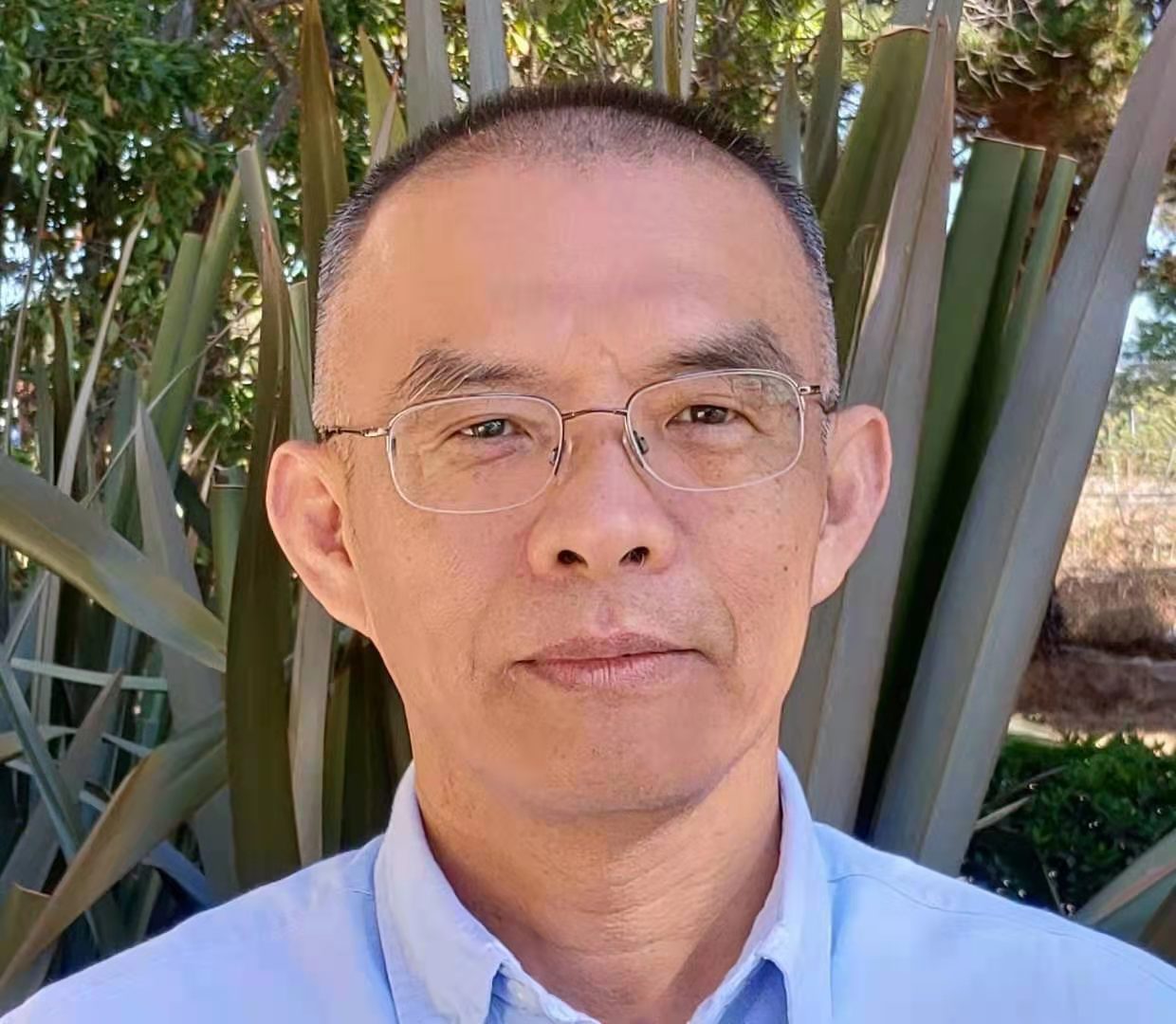}}]{Ching-yao Chan}
received his Ph.D. degree in Mechanical Engineering from University of California, Berkeley, US, in 1988. He has three decades of research experience in a broad range of automotive and transportation systems, and now is the Program Leader at California PATH and the Co-Director of Berkeley DeepDrive Consortium (bdd.berkeley.edu).He is now leading research in several topics, including driving behavior learning, pedestrian-vehicle interaction, sensor fusion for driving policy adaptation, and supervisory control in automated driving systems, human factors in automated driving.
\end{IEEEbiography}
\begin{IEEEbiography}[{\includegraphics[width=1in,height=1.25in,clip,keepaspectratio]{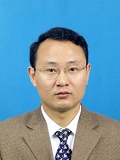}}]{Xiaofeng Liu}
received the B.S. degree in electronics engineering and the M.S. degree in computer science from the Taiyuan University of Technology in 1997 and 1999, respectively, and the Ph.D. degree in biomedical engineering from Xi'an Jiaotong University in 2006. From 2013 to 2014 he was a visiting scholar in the University College London. Since 2011 He has been a full Professor with the Hohai University, China, where he is also the Leader of the Cognition and Robotics Laboratory and the vice Director of Jiangsu Key laboratory of Special Robotic Technologies. He has over 16 grants as a PI and over 12 grants as a Researcher, including the National High-Tech R\&D Program (863) and the National Basic Research Program (973). He has been granted 15 patents and authored 20 accredited journal papers. His current research interests focus on human robot interactions, social robotics, bioinspired navigation and neural engineering.
\end{IEEEbiography}
\vspace{-35pt}
\begin{IEEEbiography}[{\includegraphics[width=1in,height=1.25in,clip,keepaspectratio]{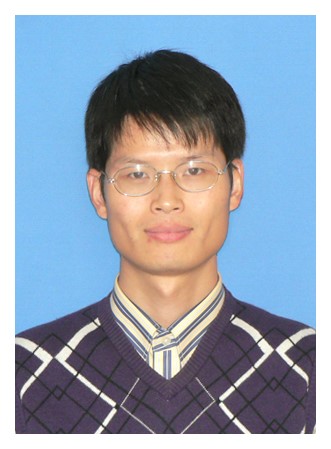}}]{Yang Chen}
received his Ph.D. degree in Harbin Engineering University, Harbi, China. He is now a associate professor at the School of Information Science and Engineering, Changzhou University. His research interests include underwater array signal processing and acoustic signal processing.
\end{IEEEbiography}






\end{document}